\documentclass[english,runningheads,a4paper]{llncs}
 
\usepackage{amssymb}
\usepackage{graphicx}
\usepackage{url}
\usepackage{tabularx}
\usepackage{chngpage}
\newcolumntype{P}[1]{>{\centering\arraybackslash}p{#1}}
\usepackage{hyperref} 
\begin{document}

\title{Weakly-Supervised Learning for Tool Localization in Laparoscopic Videos}
\titlerunning{Weakly-Supervised Learning for Tool Localization in Laparoscopic Videos}

\author{Armine Vardazaryan\inst{1}
\and Didier Mutter\inst{2}
\and Jacques Marescaux\inst{2}
\and \\Nicolas Padoy\inst{1}}

\authorrunning{Vardazaryan et al.}
\institute{ICube, University of Strasbourg, CNRS, IHU Strasbourg, France\\
\email{\{vardazaryan,npadoy\}@unistra.fr} \and University Hospital of Strasbourg, IRCAD, IHU Strasbourg, France}

\maketitle

\begin{abstract}
Surgical tool localization is an essential task for the automatic analysis of endoscopic videos. In the literature, existing methods for tool localization, tracking and segmentation require training data that is fully annotated, thereby limiting the size of the datasets that can be used and the generalization of the approaches. In this work, we propose to circumvent the lack of annotated data with weak supervision. We propose a deep architecture, trained solely on image level annotations, that can be used for both tool presence detection and localization in surgical videos. Our architecture relies on a fully convolutional neural network, trained end-to-end, enabling us to localize surgical tools without explicit spatial annotations. We demonstrate the benefits of our approach on a large public dataset, \textit{Cholec80}, which is fully annotated with binary tool presence information and of which 5 videos have been fully annotated with bounding boxes and tool centers for the evaluation.
\end{abstract}


\section{Introduction}
The automatic analysis of surgical videos is at the core of many potential assistance systems for the operating room. The localization of surgical tools, in particular, is required in many applications, such as the analysis of tool-tissue interactions, the development of novel human-robot assistance platforms and the automated annotation of video databases.

In the literature, surgical tool localization has traditionally been approached with fully supervised methods \cite{BOUGET:media17}, with the most recent localization and segmentation methods relying on deep learning \cite{garcia:iros17,jin:wacv17,kurmann:miccai17,laina:miccai17,sahu:ijcars17}. 
However, training fully supervised approaches require the data to be fully annotated with spatial information, which is tedious and expensive. 
This may explain why the datasets used so far for tool localization are small, namely in the order of a few thousand images and with a maximum of 5-6 sequences, as described in the recent review \cite{BOUGET:media17}. This then limits the applicability and generalizability of the approaches that can be developed. 

Recently, it has been shown that when a convolutional neural network is trained for the task of classification, the convolutional layers of the network learn general notions about the detected objects. 
Some recent works have used this fact to successfully localize objects in images without explicitly training for localization \cite{oquab:cvpr15,singh:iccv2017,zhou:corr14}. 
The proposed deep learning approaches directly output spatial heat maps, where the detected position corresponds to the strongest activations. 
This is achieved by replacing all fully connected layers with equivalent convolutions or removing them altogether. 
The resulting architectures are called fully convolutional networks (FCNs). 
Others have extended this approach to address the challenging task of semantic segmentation with weak supervision \cite{durand:cvpr17,kim:iccv17,saleh:pami18}. 
In the medical community as well, weakly supervised learning (WSL) has been applied to tasks such as detection of cancerous regions in medical images \cite{hwang:miccai16,jia:tmi17}. Along with the recent release of large public surgical video datasets, such as \textit{Cholec80} \cite{twinanda:tmi17}, which contains 80 complete cholecystectomy videos fully annotated with binary tool presence information ($\sim$180K frames in total), WSL techniques can potentially help develop tool localization methods that can scale up to larger datasets containing much more variability. 

In this paper, we propose a method for detecting and localizing surgical tools. It is based on weakly-supervised learning using only image-level labels and does not require any spatial annotation. Our contributions are twofold: (1) we propose the first surgical tool localization approach based on weakly-supervised learning; (2) we demonstrate our approach on the largest public endoscopic video dataset to date, namely \textit{Cholec80} \cite{twinanda:tmi17}.

\section{Methodology}
In this work, we present a method for the localization of surgical tools in endoscopic videos that does not require spatial annotations. This is possible with a FCN architecture that preserves the spatial information and permits us to observe activation regions where the tool is detected. Therefore, our method addresses two tasks: binary presence classification and tool localization, with the latter hinging on the former.

Our model takes an image as input and returns $C$ localization heat maps, where $C$ is the number of tools to be detected. For our task on the \textit{Cholec80} dataset, $C=7$. The heatmaps are used to find confidence values for each class and perform the binary classification.

\subsection{Network Architecture}
\begin{figure}[t]
    \centering
    \includegraphics[width=0.9\textwidth]{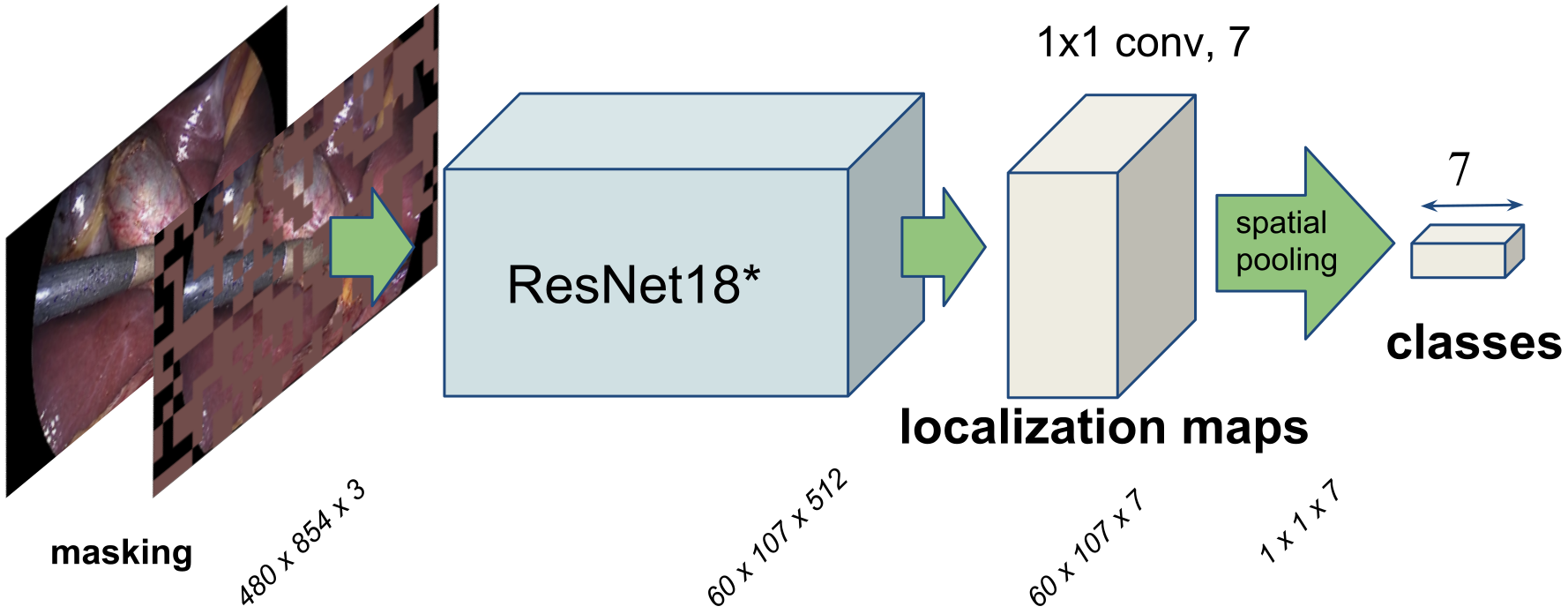}
    \caption{Our FCN for tool detection consists of a modified Resnet architecture, a $1\times 1$ convolutional layer and a spatial pooling layer.}
    \label{fig:model_fcn}
\end{figure}

As basis for our network (illustrated in Figure \ref{fig:model_fcn}), we use ResNet18 \cite{he:cvpr16} because it has been shown to perform well on a multitude of tasks. 
Since we want to preserve relative spatial information throughout our network, we remove the fully connected layer and average pooling from the end of the network. 
Additionally, we change the stride in the last two banks of ResNet from 2 to 1 pixel to obtain localization maps with a higher resolution. Note that reducing the strides for all banks would dramatically increase the dimensions of intermediate tensors during training, making it computationally infeasible.
These changes have the collective effect of quadrupling the resolution of the output. 
Using images of size $480 \times 854$ as input to the network, we obtain a feature map tensor of $60 \times 107 \times 512$ at the output of ResNet and a global stride of 8.

Then, we convert the 512 feature maps into localization maps by adding a convolutional layer of $1 \times 1$ kernels. 
To obtain one map per class, we set the number of filters in this layer to $C$.
Finally, with pooling we transform these maps into a vector of class-wise confidence values, which are, in turn, used for the binary classification of the tools. 
Instead of using conventional max pooling, we use the extended spatial pooling (ESP) $s^c=\max z^c + \alpha \min z^c$ from \cite{durand:cvpr17}, which extracts more details about the detection of the object.
In the equation, $z^c\in  {\rm I\!R}^2$ is the localization map for class $c$ and $\alpha$ is 0.6 as advised by \cite{durand:cvpr17}.

During inference, we use the raw localization maps to find the predicted position of the tools. First, the localization maps are resized to the original size of the input image with bilinear interpolation. Then, the position of the maximum activation is considered to be the predicted location of the tool.

\subsection{Training}
Before training on \textit{Cholec80}, the ResNet layers are initialized from ImageNet weights. During training, data is first randomly shuffled and batched, then data augmentation is applied independently to each image in a batch.

\subsubsection{Data Augmentation}
During training, all images in the batch are augmented before being given to the network. Augmentation includes horizontal flipping, random rotation by +90/-90 degrees, as well as the masking procedure introduced in \cite{singh:iccv2017}. Masking entails randomly replacing patches in the image with the mean pixel of the train set. This improves the quality of predicted localization maps.
\subsubsection{Loss}
The models are trained for multi-label classification with a weighted cross-entropy loss $\mathcal{L}$ presented in Equation \ref{eq:loss}, where $k_c$ and $v_c$ are respectively the ground truth and predicted tool presence for class $c$, $\sigma$ is the sigmoid function, and $\mathcal{W}_c$ is the weight for class $c$. Weights are added to counteract the polarizing effect of class imbalance. The weight for each class is inversely proportional to the number of occurrences of the class in the train set.

\begin{equation}\label{eq:loss}
\mathcal{L} = \sum_{c=1}^C\frac{-1}{N}\left [ \mathcal{W}_c k_c \log(\sigma(v_c)) + (1-k_c) \log(1-\sigma(v_c)) \right]
\end{equation}

\section{Experimental Results}
\subsection{Setup}
For our experiments, we use the \textit{Cholec80} dataset \cite{twinanda:tmi17} containing 80 videos of cholecystectomy procedures, fully annotated with image-level surgical tool labels for binary detection. Our training, validation and test sets consist of 40, 10 and 30 videos, respectively. Additionally, for the purpose of evaluating the performance of our localization method, our team has fully annotated 5 videos from the test set with bounding boxes and tool centers. The details of these annotations are presented in Table \ref{annotations_details}. They are also illustrated in column 1 of Figure \ref{fig:qual_all}. As part of the preprocessing, we randomly mask patches of $30\times 30$ by filling these squares with the average pixel value of the test set. For each patch, the probability of masking is 0.5.
We train all the evaluated models for 120 epochs with an initial learning rate of 0.1, which decreases by a factor of 10 at [60, 100] epochs. That learning rate is applied to the new convolutional layer, while the layers of ResNet are trained with a learning rate smaller by a factor of 100. 
In our loss function, we use a weight decay of $10^{-4}$.
The models were trained with the momentum optimizer (momentum $\mu=0.9$) and batch size of 16.
\begin{table}[t]
\centering
\begin{tabular}{|P{1.4cm}|P{1.2cm}|P{1.3cm}|P{1cm}|P{1.3cm}|P{1.3cm}|P{1.3cm}|P{1.3cm}|P{1cm}|}
\hline
  & Grasper & Bipolar & Hook & Scissors & Clipper & Irrigator & Spec.Bag & Total\\ \hline
 Sp.Annot.& 4774    & 379     & 4313 & 327      & 384     & 332       & 375 & 7175 \\ \hline
 Cholec80 & 102588 & 8876 & 103106 & 3254 & 5986 & 9814 & 11462 & 161397\\ \hline
\end{tabular}
\smallskip
\caption{Dataset statistics. (Row 1) Number of frames where each tool is present, for the 5 spatially annotated videos. (Row 2) Number of frames where each tool is present in the complete \textit{Cholec80} dataset.}
\label{annotations_details}
\end{table}

\subsection{Evaluated Models}
We evaluate several variants of the architecture presented in section 2.1 in order to compare the differences and search for the best performing configuration. The models we devised are as follows: FCN\_ESP (M1), FCN\_ESP\_Msk (M2), FCN\_ESP\_MM (M3), FCN\_ESP\_MM\_Msk (M4), FCN\_MSP (M5), FCN\_MSP\_Msk (M6), FCN\_MSP\_MM (M7), FCN\_MSP\_MM\_Msk (M8). The models M1-M4 use the ESP method seen in section 2.1. To see whether that spatial pooling method is beneficial, we included identical models that use max pooling (MSP) instead: M5-M8.
Similarly, to evaluate the benefit of masking images during training, architectures M2, M4, M6 and M8 incorporate masking, while M1, M3, M5 and M7 do not. Finally, models M3, M4, M7 and M8 use multi-maps \cite{durand:cvpr17}, described below.

\subsubsection{Multi-maps}
Our network architecture contains a convolutional layer of 7 kernels, each dedicated to one tool. Introduced in \cite{durand:cvpr17}, the notion of multi-maps is based on the following idea: instead of using a single kernel for each class, multiple kernels can be used and be followed by class-wise averaging to obtain 7 localization maps. 
This helps the network to extract more details about the object than when a single feature map is used.
The authors of \cite{durand:cvpr17} advise to use 8 kernels per class. However, since the objects we detect are significantly simpler than the classes used in \cite{durand:cvpr17}, we use only 4 kernels per class (28 filters altogether).

\subsection{Classification}
\begin{figure}[t]
    \centering
    \includegraphics[width=0.7\textwidth]{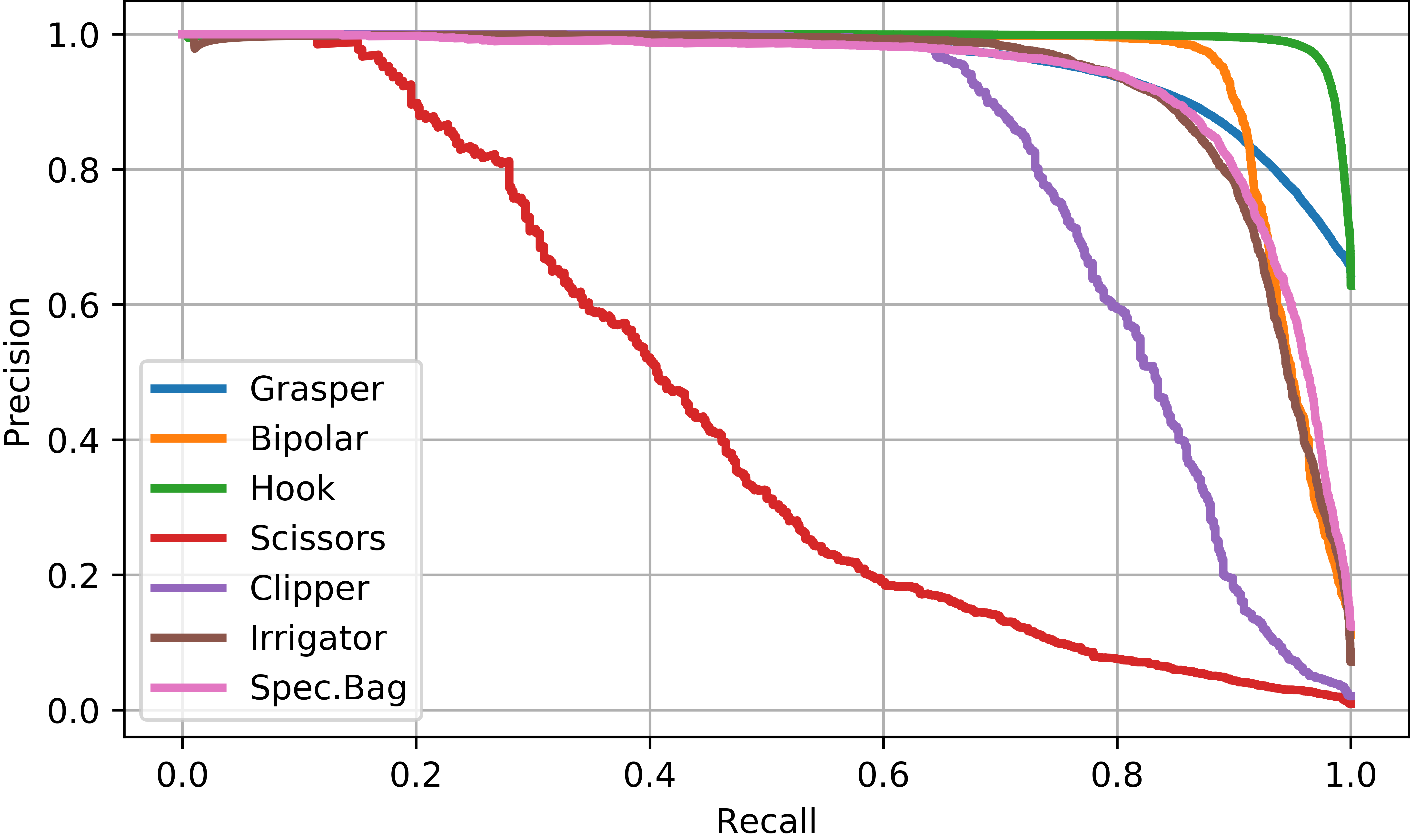}
    \caption{Precision-recall classification curve for FCN\_ESP\_MM\_Msk  (best seen in color).}
    \label{fig:class_ap}
\end{figure}

As mentioned above, we use the dataset \textit{Cholec80} to test our method. Specifically, the 30 videos of the test set are used for testing the classification performance. 
To quantify the results, we use average precision (AP), which is defined as the area under the precision-recall curve. We illustrate the curve for architecture FCN\_ESP\_MM\_Msk in Figure \ref{fig:class_ap}, where we see that results for scissors and clipper fall behind the rest of the tools.
A similar pattern can be observed in Table \ref{tab:class_ap}. All models detect most tools quite well with AP values above 93\%. However, the results for scissors ($\sim$50\%) and clipper ($\sim$82\%) are significantly worse than those of the other tools. This may be due to the fact that scissors and clipper are present only in 2\% and 4\% of annotations, respectively. In contrast, hook is present in 64\% of all annotations (see Table \ref{annotations_details}, row 2).
\begin{table}[t]
\centering
\begin{tabular}{|l|c|c|c|c|c|c|c|c|}
\hline
             & Grasper & Bipolar & Hook & Scissors & Clipper & Irrigator & Spec.bag & \textbf{mAP} \\ \hline
FCN\_ESP           & 96.5          & 94.9          & 99.5          & \textbf{51.4} & 81.4          & 93.2          & 93.7          & 87.2 \\ \hline
FCN\_ESP\_Msk     & 96.7          & \textbf{95.5} & \textbf{99.6} & 50.0          & 82.3          & \textbf{94.3} & 93.5          & \textbf{87.4} \\ \hline
FCN\_ESP\_MM       & 96.6          & 95.0          & \textbf{99.6} & 50.2          & 82.8          & 94.0          & 93.5          & \textbf{87.4} \\ \hline
FCN\_ESP\_MM\_Msk & 96.7          & 94.8          & 99.5          & 44.2          & 81.9          & 92.9          & 93.2          & 86.1 \\ \hline
FCN\_MSP           & 96.6          & 93.9          & 99.5          & 49.6          & 81.6          & 92.1          & 92.4          & 86.5 \\ \hline
FCN\_MSP\_Msk     & 96.7          & 94.1          & 99.5          & 49.4          & \textbf{83.2} & 92.7          & 93.4          & 87.0 \\ \hline
FCN\_MSP\_MM       & 96.7          & 93.8          & 99.5          & 50.4          & 81.8          & 91.5          & 92.7          & 86.6 \\ \hline
FCN\_MSP\_MM\_Msk & \textbf{96.8} & 94.2          & \textbf{99.6} & 49.8          & 83.0          & 93.3          & \textbf{94.0} & 87.2 \\ \hline
\end{tabular}
\smallskip
\caption{Average precision (AP) for binary tool presence classification.}
\label{tab:class_ap}
\end{table}

\subsection{Localization}
With our method, we are able to obtain localization maps that contain information about the positions of the tools in the frame. Multiple classes of tools can be detected in the same frame. Note, however, that our approach is not designed to detect multiple instances of the same class, because all instances would share the same localization map. In this work, we limit detection to a single instance of each type of tool, even though multiple instance detection could, for example, be possible with post-processing heuristics.

We evaluate the quality of the predictions by comparing them against the ground truth bounding boxes that we have annotated for that purpose. In the cases where multiple instances of the same tool are present in the frame, we pick the bounding box closest to the prediction.

\subsubsection{Localization AP}
To evaluate the quality of localization, we compute AP as described in \cite{hwang:miccai16}, which is based on a metric defined in \cite{oquab:cvpr15}. 
If the predicted location lies in a ground truth bounding box of the same class, with a tolerance of 8 pixels (the global stride of the network), the example is considered a true positive. Otherwise, it is a false positive. Taking that into account, we compute precision and recall as described in \cite{everingham:ijcv10}, where recall is defined as the proportion of positive predictions, and precision is the proportion of true positives in positive predictions. AP is then computed as the area under the precision-recall curve. For this evaluation, we use only the positive classes as the negative class corresponds to having no tool in the image and cannot be annotated with a bounding box.
The results of this computation are presented in Table \ref{tab:loc_ap}. The localization AP values for all models are similar, ranging approximately between 87\% and 89\%. Our intuition is that all models are almost equally likely to predict a tool center that lies in the bounding box, without capturing the quality of the precise location inside the bounding box. In the next section, we quantify the accuracy of the predicted tool centers relative to ground truth.
\begin{table}[t]
\centering
\begin{tabular}{|l|c|c|c|c|c|c|c|c|}
\hline
             & Grasper & Bipolar & Hook & Scissors & Clipper & Irrigator & Spec.bag & \textbf{mAP} \\ \hline
FCN\_ESP & \textbf{97.9}  & 99.3 & 98.6 & 63.9 & 97.5 & 94.2 & 69.5 & 88.7 \\ \hline
FCN\_ESP\_Msk & 96.6  & 99.6 & \textbf{99.0} & 52.6 & \textbf{98.8} & 93.3 & 72.1 & 87.4 \\ \hline
FCN\_ESP\_MM & 97.1  & \textbf{99.7} & 98.9 & 64.8 & 98.3 & 88.7 & 72.1 & 88.5 \\ \hline
FCN\_ESP\_MM\_Msk   & 96.9  & 99.5 & 97.9 & 58.1 & 97.9 & 91.8 & \textbf{78.4} & 88.7 \\ \hline
FCN\_MSP     & 97.4 & 99.6 & 98.0 & 57.4 & 98.2 & 94.3 & 70.7 & 88.0 \\ \hline
FCN\_MSP\_Msk & 96.5 & 99.5 & 98.7 & \textbf{66.4} & 98.4 & 94.3 & 67.5 & \textbf{88.8} \\ \hline
FCN\_MSP\_MM & \textbf{97.9} & \textbf{99.7} & 98.7 & 46.3 & 97.7 & \textbf{94.4} & 73.5 & 86.9 \\ \hline
FCN\_MSP\_MM\_Msk & 97.6 & 99.5 & 98.9 & 57.9 & 97.2 & 92.9 & 72.7 & 88.1 \\ \hline
\end{tabular}
\smallskip
\caption{Localization average precision (AP) for the 8 evaluated models.}
\label{tab:loc_ap}
\end{table}

\subsubsection{Distance Error}
Localization AP gives a coarse idea about the quality of obtained predictions. To get a better sense of the accuracy of the localization, we compute the distance between the predicted tool center and its ground truth. We normalize this value by the diagonal of the image. The results are presented in Table \ref{tab:dist}. We can see that, generally, masking and ESP improve the quality of predicted tool centers. On the other hand, multi-maps do not seem to affect the outcome significantly. It is also noteworthy that specimen bag is localized significantly worse than the other tools. This can be explained by the varying shape of the bag, as well as the ambiguity of its center.
\begin{table}[t]
\centering
\begin{tabular}{|l|c|c|c|c|c|c|c|c|}
\hline
             & Grasper & Bipolar & Hook & Scissors & Clipper & Irrigator & Spec.bag & \textbf{mean} \\ \hline
FCN\_ESP & \textbf{6.7} & 5.4 & 6.0 & 12.0 &  8.4 & 4.4 &  9.7 & 7.5 \\ \hline
FCN\_ESP\_Msk & \textbf{6.7} & 4.6 & \textbf{4.7} & 11.3 &  \textbf{6.7} & 4.7 &  \textbf{9.1} & \textbf{6.8} \\ \hline
FCN\_ESP\_MM  & 6.8 & 5.0 & 5.5 & 11.0 &  8.3 & 4.5 &  \textbf{9.1} & 7.1 \\ \hline
FCN\_ESP\_MM\_Msk & 6.9 & 5.3 & 5.7 & \textbf{ 9.6} &  6.9 & 4.6 &  \textbf{9.1} & 6.9 \\ \hline
FCN\_MSP     & 7.4 & \textbf{4.4} & 5.9 & 17.8 & 8.7 & \textbf{4.0} & 10.4 & 8.4 \\ \hline
FCN\_MSP\_Msk & 7.1 & 4.6 & 5.3 & 10.1 & 7.4 & 4.2 & 9.4 & 6.9 \\ \hline
FCN\_MSP\_MM     & 7.6 & 4.8 & 5.8 & 17.7 & 9.2 & 3.9 & 10.0 & 8.4 \\ \hline
FCN\_MSP\_MM\_Msk & \textbf{6.7} & 4.7 & 5.3 & 11.6 & 7.9 & 4.3 & 9.2 & 7.1 \\ \hline
\end{tabular}
\smallskip
\caption{Mean distance from predicted tool center to true center in percents (relative to the image diagonal).}
\label{tab:dist}
\end{table}

\subsubsection{Qualitative Results}
\begin{figure}[t]
    \centering
    \includegraphics[width=\textwidth]{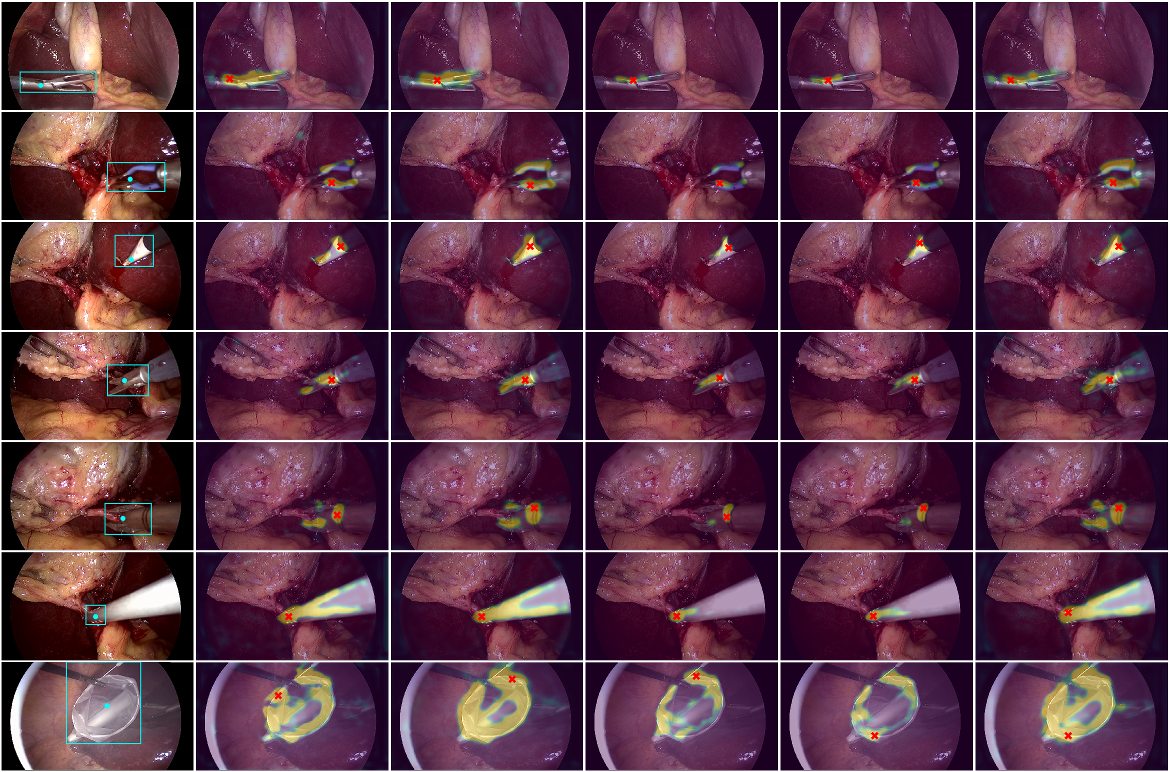}
    \caption{Column 1: Ground truth bounding box and tool center. Columns 2-6: input images overlaid with corresponding localization maps (after sigmoid) and predicted tool centers for FCN\_ESP, FCN\_ESP\_Msk, FCN\_MSP\_Msk, FCN\_MSP\_MM\_Msk, FCN\_ESP\_MM\_Msk, in that order.}
    \label{fig:qual_all}
\end{figure}

For the sake of visual comparison, we present qualitative results for 8 evaluated models in Figure \ref{fig:qual_all}, where input images are overlaid with localization maps. Just as the quantitative results suggest, the performances of the networks are very similar and the detected tool centers are very close to one another in most cases. However, the models with masking and ESP generate more detailed maps that cover the tools better than other models and provide strong ROI for the tools.

\begin{figure}[t]
    \centering
    \includegraphics[width=\textwidth]{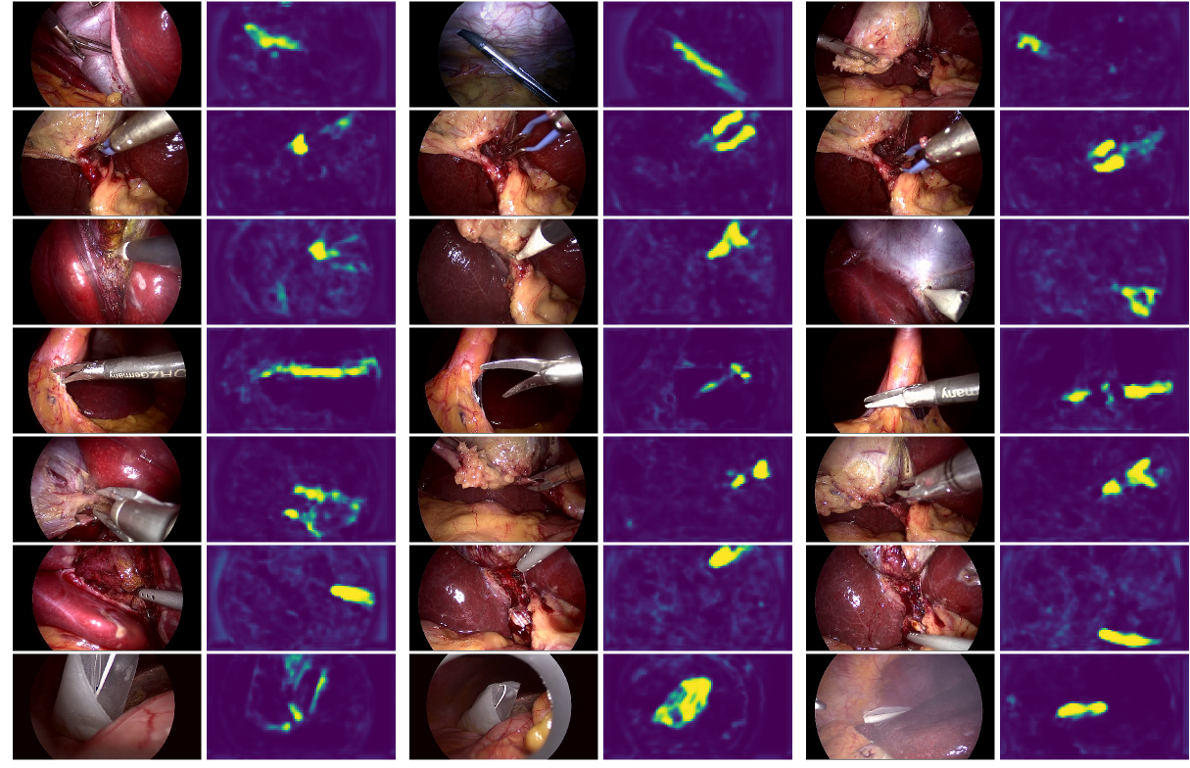}
    \caption{Input images (left) and corresponding localization maps after sigmoid layer (right). Each row shows 3 examples of the same tool. The tools in rows 1-7 are presented in the following order: grasper, bipolar, hook, scissors, clipper, irrigator, specimen bag. These results correspond to architecture FCN\_ESP\_MM\_Msk. (Best seen in color)}
    \label{fig:score_maps}
\end{figure}

In Figure \ref{fig:score_maps}, we present additional results for the architecture FCN\_ESP\_MM\_Msk.
In the figure, we see which features the network finds most discriminative about each of the tools. Ideally, we aim to localize the working end of the tools only, as the shaft does not usually contain tool-specific features. In Figure \ref{fig:score_maps}, we can see that for scissors and irrigator (row 4 and 6 respectively) the shafts themselves are very distinctive and discriminative. In the case of scissors, the brightest detection corresponds to the shaft. This may explain why the localization AP values for scissors are the lowest among all tools, as the annotated bounding boxes are defined over tool tips only (see column 1 in Figure \ref{fig:qual_all}). Specimen bag (last row) is an exception since it is not connected to a shaft.
We should also note that the second tool, bipolar, is not fully detected. The network detects the blue insulated section of the forceps but not the metal tips. Our intuition is that they look very similar to those of grasper and hence cannot be used to discriminate one tool from the other. Additional qualitative results can be seen in the supplementary video (\href{https://youtu.be/7VWVY04Z0MA}{https://youtu.be/7VWVY04Z0MA}).

\section{Conclusions}
In this work, we showed that reliable surgical tool detection and localization can be achieved without the use of spatial annotations during training. Our method relies on a FCN architecture that preserves relative spatial information of the input image. This enables us to localize the surgical tools while using only binary presence annotations for training. We evaluated several variants of our network, obtaining very promising AP values of around 87 and 88 for classification and localization on the test set, respectively. 
These results also suggest that the proposed approach could be used to ease the generation of spatial annotations within surgical video labeling software and extended for tool segmentation.

\subsubsection{Acknowledgements.}
This work was supported by French state funds managed within the Investissements d'Avenir program by BPI France (project CONDOR) and by the ANR (references ANR-11-LABX-0004 and ANR-10-IAHU-02). The authors would also like to acknowledge the support of NVIDIA with the donation of a GPU used in this research.

\bibliographystyle{splncs04}
\bibliography{tool_localization}
\end{document}